\documentclass[twoside,leqno,twocolumn]{article}  
\usepackage{ltexpprt} 
\usepackage{times}
\usepackage{color}
\usepackage{graphicx}
\usepackage{amsmath}
\usepackage{algorithm} 
\usepackage{algorithmic}
\usepackage{subfigure}
\usepackage{url}
\usepackage{bm}
\usepackage{bbm}
\usepackage{multirow}

\begin{document}

\title{\Large Learning A Task-Specific Deep Architecture For Clustering}
\author{Zhangyang Wang\thanks{Beckman Institute, University of Illinois at Urbana-Champaign, IL 61821. \{zwang119, chang87, t-huang1\}@illinois.edu.} \\
\and 
Shiyu Chang$^*$
\and 
Jiayu Zhou\thanks{Department of Computer Science and Engineering, Michigan State University, MI 48824. jiayuz@msu.edu}
\and 
Meng Wang\thanks{Hefei University of Technology, Hefei, Anhui, 230009, China. wangmeng@hfut.edu.cn.}
\and 
Thomas S. Huang$^*$
}
\date{}

\maketitle


\begin{abstract} \small\baselineskip=9pt 
While sparse coding-based clustering methods have shown to be successful, their bottlenecks in both efficiency and scalability limit the practical usage. In recent years, deep learning has been proved to be a highly effective, efficient and scalable feature learning tool. In this paper, we propose to emulate the sparse coding-based clustering pipeline in the context of deep learning, leading to a carefully crafted deep model benefiting from both. A feed-forward network structure, named TAGnet, is constructed based on a graph-regularized sparse coding algorithm. It is then trained with task-specific loss functions from end to end. We discover that connecting deep learning to sparse coding benefits not only the model performance, but also its initialization and interpretation. Moreover, by introducing auxiliary clustering tasks to the intermediate feature hierarchy, we formulate DTAGnet and obtain a further performance boost. Extensive experiments demonstrate that the proposed model gains remarkable margins over several state-of-the-art methods. 
\end{abstract}


\section{Introduction}
Clustering aims to learn the hidden data patterns and group similar structures in a unsupervised way. While many classical clustering algorithms have been proposed, such as K-means, Gaussian mixture model (GMM) clustering \cite{biernacki2000assessing}, maximum-margin clustering \cite{xu2004maximum} and information theoretic clustering \cite{li2004minimum}, most only work well when the data dimensionality is low. Since high-dimensional data exhibits dense grouping in low-dimensional embeddings \cite{nie2009spectral}, researchers have been motivated to  first project the original data into a low-dimensional subspace \cite{roth2003feature} and then clustering on the feature embeddings. Among many feature embedding learning methods, sparse codes \cite{wright2009robust} are proven to be robust and efficient features for clustering, as verified by many  \cite{cheng2010learning, zheng2011graph}. 


Effectiveness and scalability are two major concerns in designing a clustering algorithm under Big Data scenarios \cite{KDD}. Conventional sparse coding models rely on iterative approximation algorithms, whose inherently sequential structure as well as the data-dependent complexity and latency often constitute a major bottleneck in the computational efficiency \cite{LISTA}. That also results in the difficulty when one tries to jointly optimize the unsupervised feature learning and the supervised task-driven steps \cite{mairal2012task}. Such a joint optimization usually has to rely on solving complex bi-level optimization \cite{bertsekas1999nonlinear}, such as \cite{IJCAI}, which constitutes another efficiency bottleneck. What is more, to effectively model and represent datasets of growing sizes, sparse coding needs to refer to larger dictionaries \cite{lee2006efficient}. Since the inference complexity of sparse coding increases more than linearly with respect to the dictionary size \cite{IJCAI}, the scalability of sparse coding-based clustering work turns out to be quite limited.

To conquer those limitations, we are motivated to introduce the tool of deep learning in clustering, to which there has been a lack of attention paid. The advantages of deep learning are achieved by its large learning capacity, the linear scalability with the aid of stochastic gradient descent (SGD), and the low inference complexity \cite{bengio2009learning}. The feed-forward networks could be naturally tuned jointly with task-driven loss functions. On the other hand, generic deep architectures \cite{ImageNet} largely ignore the problem-specific formulations and prior knowledge. As a result, one may encounter difficulties in choosing optimal architectures, interpreting their working mechanisms, and initializing the parameters.

\begin{figure*}[htbp]
\centering
\begin{minipage}{0.60\textwidth}
\centering \subfigure[] {
\includegraphics[width=\textwidth]{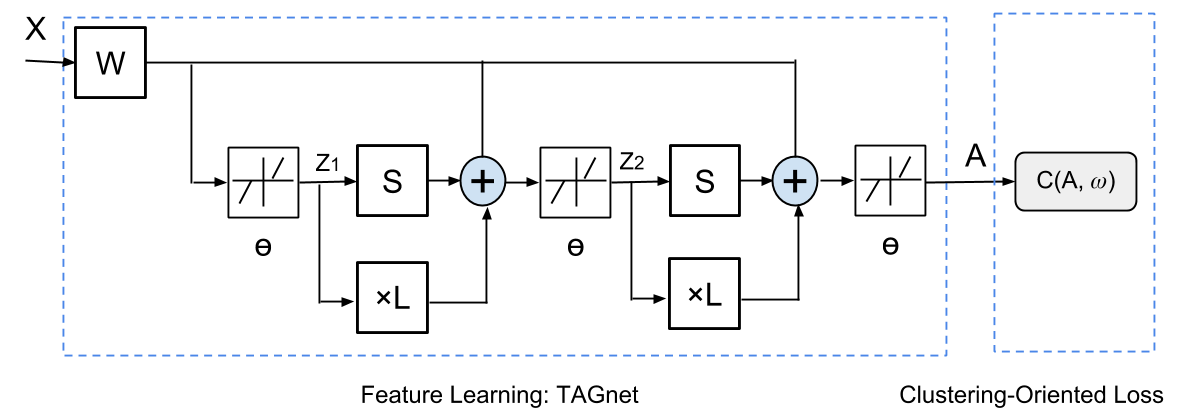}
}\end{minipage}
\begin{minipage}{0.24\textwidth}
\centering \subfigure [] {
\includegraphics[width=\textwidth]{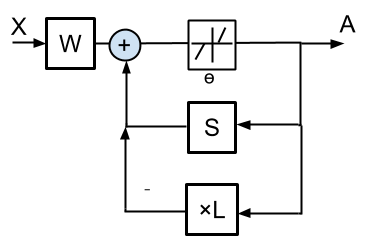}
}\end{minipage}
\caption{(a) The proposed pipeline, consisting of the TAGnet network for feature learning, followed by the clustering-oriented loss functions. The parameters $\mathbf{W, S}, \bm{\theta}$ and $\bm{\omega}$ are all learnt end-to-end from training data. (b) The block diagram of solving (\ref{eqnISTA}).}
\label{figISTA}
\end{figure*}


In this paper, we demonstrate how to \textbf{combine the sparse coding-based pipeline into deep learning models for clustering}. The proposed framework takes advantage of both sparse coding and deep learning. Specifically, the feature learning layers are inspired by the graph-regularized sparse coding inference process, via reformulating iterative algorithms \cite{LISTA} into a feed-forward network, named \textbf{TAGnet}. Those layers are then jointly optimized with the task-specific loss functions from end to end. Our technical novelty and merits are summarized in three-folds:
\begin{itemize}
\item As a deep feed-forward model, the proposed framework provides extremely efficient inference process and high scalability to large scale data. It allows to learn more descriptive features than conventional sparse codes. 
\item  We discover that incorporating the expertise of sparse code-based clustering pipelines \cite{cheng2010learning, zheng2011graph} improves our performances significantly. Moreover, it greatly facilitates the model initialization and interpretation. 
\item  We further enforce auxiliary clustering tasks on the hierarchy of features, we develop \textbf{DTAGnet} and observe further performance boosts on the CMU MultiPIE dataset \cite{MultiPIE}. 
\end{itemize}
 
\section{Related Work}
\subsection{Sparse coding for clustering}
Assuming data samples $\mathbf{X }= [\mathbf{x}_1, \mathbf{x}_2,\cdots, \mathbf{x}_n]$, where $\mathbf{x}_i \in \mathbf{R}^{m \times 1}$ and $i= 1,2, \cdots, n$. They are encoded into sparse codes $\mathbf{A}= [\mathbf{a}_1, \mathbf{a}_2,\cdots, \mathbf{a}_n]$, where $\mathbf{a}_i \in \mathbf{R}^{p \times 1}$ and $i=1, 2, \cdots, n$, using a learned dictionary $\mathbf{D}= [\mathbf{d}_1, \mathbf{d}_2,\cdots, \mathbf{d}_p]$, where $ \mathbf{d}_i \in \mathbf{R}^{m \times 1}, i=1,2, \cdots, p$ are the learned atoms. The sparse codes are obtained by solving the following convex optimization ($\lambda$ is a constant): 
\begin{equation}
\begin{array}{l}\label{e000}
\mathbf{A} = \arg \min_{\mathbf{A}} \frac{1}{2}||\mathbf{X - D A}||_F^2 + \lambda\sum_i ||\mathbf{a}_i||_1,
\end{array}
\end{equation}
In \cite{cheng2010learning}, the authors suggested that the sparse codes can be used to construct the similarity graph for spectral clustering \cite{ng2002spectral}. Furthermore, to capture the geometric structure of local data manifolds, the graph regularized sparse codes are further suggested in \cite{zheng2011graph, yang2014data} by solving: 
\begin{equation}
\begin{array}{l}\label{e00}
\mathbf{A} = \arg \min_{\mathbf{A}} \frac{1}{2}||\mathbf{X - D A}||_F^2 + \lambda\sum_i ||\mathbf{a}_i||_1\\ \qquad + \frac{\alpha}{2}Tr (\mathbf{ALA^T}),
\end{array}
\end{equation}
where $\mathbf{L}$ is the graph Laplacian matrix and can be constructed from a pre-chosen pairwise similarity (affinity) matrix $\mathbf{P}$. More recently in \cite{IJCAI}, the authors suggested to simultaneously learn feature extraction and discriminative clustering, by formulating a task-driven sparse coding model \cite{mairal2012task}. They proved that such joint methods consistently outperformed non-joint counterparts. 

\subsection{Deep learning for clustering}
In \cite{tian2014learning}, the authors explored the possibility of employing deep learning in graph clustering. They first learned a nonlinear embedding of the original graph by an auto encoder (AE),
followed by a K-means algorithm on the embedding to obtain the final clustering result. However, it neither exploits more adapted deep architectures nor performs any task-specific joint optimization. In \cite{chen2015deep}, a deep belief network (DBN) \cite{hinton2006fast} with nonparametric clustering was presented. As a generative graphical model, DBN provides a faster feature learning, but is less effective than AEs in terms of learning discriminative features for clustering. In \cite{JMLR}, the authors extended the semi non-negative matrix factorization (Semi-NMF) model \cite{li2006relationships} to a Deep Semi-NMF model, whose architecture resembles stacked AEs. Our proposed model is substantially different from all these previous approaches, due to its unique task-specific architecture derived from sparse coding domain expertise, as well as the joint optimization with clustering-oriented loss functions.

\section{Model Formulation}
The proposed pipeline consists of two blocks. As depicted in Fig. \ref{figISTA} (a), it is trained end-to-end in an unsupervised way. It includes a feed-forward architecture, termed \textit{Task-specific And Graph-regularized Network} (\textbf{TAGnet}), to learn discriminative features, and the clustering-oriented loss function. 

\subsection{TAGnet: Task-specific And Graph-regularized Network}


Different from generic deep architectures, TAGnet is designed in a way to take advantage of the successful sparse code-based clustering pipelines \cite{zheng2011graph, IJCAI}. It aims to learn features that are optimized under clustering criteria, while encoding graph constraints (\ref{e00}) to regularize the target solution. TAGnet is derived from the following theorem:
\begin{theorem} \label{E}
The optimal sparse code $\mathbf{A}$ from (\ref{e00}) is the fixed point of 
\begin{equation}
\begin{array}{l}\label{eqnISTA}
\mathbf{A} = h_{\frac{\lambda}{N}}[(\mathbf{I} - \frac{1}{N}\mathbf{D}^T\mathbf{D}) \mathbf{A} - \mathbf{A} (\frac{\alpha}{N} \mathbf{L}) + \frac{1}{N}\mathbf{D}^T\mathbf{X}],
\end{array}
\end{equation}
where $h_{\bm{\theta}}$ is an element-wise shrinkage function parameterized by $\bm{\theta}$:
\begin{equation}
\begin{array}{l}\label{threshold}
[h_{\bm{\theta}}(\mathbf{u})]_i = \text{sign}(\mathbf{u}_i)(|\mathbf{u}_i| - \bm{\theta}_i)_{+}.
\end{array}
\end{equation}
$N$ is an upper bound on the largest eigenvalue of $\mathbf{D}^T\mathbf{D}$. 
\end{theorem}
The complete proof of Theorem \ref{E} can be found in the supplementary. Theorem \ref{E} outlines an iterative algorithm to solve (\ref{e00}). Under quite mild conditions \cite{FISTA}, after $\mathbf{A}$ is initialized, one may repeat the shrinkage and thresholding process in (\ref{eqnISTA}) until convergence. Moreover, the iterative algorithm could be alternatively expressed as the block diagram in Fig. \ref{figISTA} (b), where 
\begin{equation}
\begin{array}{l}\label{scpara}
\mathbf{W}$ = $\frac{1}{N}\mathbf{D}^T, \mathbf{S} = \mathbf{I} - \frac{1}{N}\mathbf{D}^T\mathbf{D}, \bm{\theta}= \frac{\lambda}{N}. 
\end{array}
\end{equation}
In particular, we define the new operator ``$\times \mathbf{L}$'':  $\mathbf{A} \rightarrow  - \frac{\alpha}{N}\mathbf{A}\mathbf{L}$, where the input $\mathbf{A}$ is multiplied by the pre-fixed $\mathbf{L}$ from the right side and scaled by the constant $- \frac{\alpha}{N}$.


By time-unfolding and truncating Fig. \ref{figISTA} (b) to a fixed number of $K$ iterations ($K$ = 2 by default)\footnote{We test larger $K$ values (3 or 4), but they do not bring noticeable performance improvements in our clustering cases.}, we obtain the TAGnet form in Fig. \ref{figISTA} (a). $\mathbf{W}$,  $\mathbf{S}$ and $\bm{\theta}$ are all to be learnt jointly from data. $\mathbf{S}$ and $\bm{\theta}$ are tied weights for both stages\footnote{Out of curiosity, we have also tried the architecture that treat $\mathbf{W}$, $\mathbf{S}$ and $\bm{\theta}$ in both stages as independent variables. We find that sharing parameters improves the performance.}. It is important to note that the output $\mathbf{A}$ of TAGnet is not necessarily identical to the predicted sparse codes by solving (\ref{e00}). Instead, the goal of TAGnet is to learn discriminative embedding that is optimal for clustering.

To facilitate training, we further rewrite (\ref{threshold}) as:
\begin{equation}
\begin{array}{l}\label{threshold1}
[h_{\bm{\theta}}(\mathbf{u})]_i = \bm{\theta}_i \cdot \text{sign}(\mathbf{u}_i)(|\mathbf{u}_i|/\bm{\theta}_i - 1)_{+} = \bm{\theta}_i h_1(\mathbf{u}_i/\bm{\theta}_i)
\end{array}
\end{equation}
Eqn. (\ref{threshold1}) indicates that the original neuron with trainable thresholds can be decomposed into two linear scaling layers plus a unit-threshold neuron. The weights of the two scaling layers are diagonal matrices defined by $\bm{\theta}$ and its element-wise reciprocal, respectively.

A notable component in TAGnet is the \textbf{$\times \mathbf{L}$ branch} of each stage. The graph laplacian $\mathbf{L}$ could be computed in advance. In the feed-forward process, a $\times \mathbf{L}$ branch takes the intermediate $\mathbf{Z}_k$ ($k$ = 1, 2) as the input, and applies the ``$\times \mathbf{L}$'' operator defined above. The output is aggregated with the output from the learnable $\mathbf{S}$ layer. In the back propagation, $\mathbf{L}$ will not be altered. In such a way, the graph regularization is effectively encoded in the TAGnet structure as a prior.

An appealing highlight of (D)TAGnet lies in its very effective and straightforward initialization strategy. With sufficient data, many latest deep networks train well with random initializations without pre-training. However, it has been discovered that poor initializations hamper the effectiveness of first-order methods (e.g., SGD) in certain cases \cite{sutskever2013importance}. For (D)TAGnet, it is however much easier to initialize the model in the right regime. That benefits from the analytical relationships between sparse coding and network hyperparameters defined in (\ref{scpara}): we could initialize deep models from corresponding sparse coding components, the latter of which is easier to obtain. Such an advantage becomes much more important when the training data is limited




\subsection{Clustering-oriented loss functions}
Assuming $K$ clusters, and $\bm{\omega} = [\bm{\omega}_1,..., \bm{\omega}_K]$ as the set of parameters of the loss function, where $\bm{\omega}_i$ corresponds to the $i$-th cluster, $i$ = $1, 2, ..., K$. In this paper, we adopt the following two forms of clustering-oriented loss functions.

One natural choice of the loss function is extended from the popular softmax loss, and take the entropy-like form as:
\begin{equation}
\begin{array}{l}\label{logistic}
C(\mathbf{A}, \bm{\omega})=  - \sum_{i=1}^{n}  \sum_{j=1}^K p_{ij}  \log p_{ij}.
\end{array}
\end{equation}
where $p_{ij}$ denotes the the probability that sample $\mathbf{x_i} $ belongs to cluster $j$, $i= 1,2, \cdots, N$ and $j= 1,2, \cdots, K$:
\begin{equation}
\begin{array}{l}\label{3}
p_{ij}=p(j|\bm{\omega, a}_i)=\frac{e^{- \bm{\omega}_j^T\mathbf{a}_i}}{\sum_{l=1}^K e^{- \bm{\omega}_l^T\mathbf{a}_i}},
\end{array}
\end{equation}
In testing, the predicted cluster label of input $\mathbf{a}_i$ is determined using the maximum likelihood criteria based on the predicted $p_{ij}$.

The maximum margin clustering (MMC) approach was proposed in \cite{xu2004maximum}. MMC finds a way to label the samples by running an SVM implicitly, and the SVM margin obtained would be maximized over all possible labels~\cite{zhao2008efficient}. By referring to the MMC definition, the authors of \cite{IJCAI} designed the max-margin loss:
\begin{equation}
\begin{array}{l}\label{hingeall}
C(\mathbf{A}, \bm{\omega})=   \frac{\lambda}{2}||\bm{\omega}||^2 + \sum_{i=1}^{n}  C(\mathbf{a}_i, \bm{\omega}). 
\end{array}
\end{equation}
In the above equation, the loss for an individual sample $\mathbf{a}_i$ is defined as:
\begin{equation}
\begin{array}{l}\label{hinge}
C(\mathbf{a}_i, \bm{\omega})= \max(0, 1+ f^{r_i}(\mathbf{a}_i) - f^{y_i}(\mathbf{a}_i))\\
\text{where} \quad y_i = \quad\underset{j=1, ..., K}{\arg\max}\quad f^{j}(\mathbf{a}_i)\\
\qquad \quad $ $  r_i = \underset{j=1, ..., K, j \neq y_i}{\arg\max} f^{j}(\mathbf{a}_i).
\end{array}
\end{equation} 
where $f^{j}$ is the prototype for the $j$-th cluster. In testing, the predicted cluster label of input $\mathbf{a}_i$ is determined by weight vector that achieves the maximum $\bm{\omega}_j^T\mathbf{a}_i$.

\noindent \textbf{Model Complexity} The proposed framework can handle large-scale and high-dimensional data effectively via the stochastic gradient descent (SGD) algorithm. In each step, the back propagation procedure requires only operations of order O($p$)~\cite{LISTA}. The training algorithm takes O($Cnp$) time ($C$ is a constant in terms of the total numbers of epochs, stage numbers, etc.). In addition, SGD is easy to be parallelized and thus could be efficiently trained using GPUs.


\subsection{Connections to Existing Models}

There is a close connection between sparse coding and neural network. In \cite{LISTA}, a feed-forward neural network, named LISTA, is proposed to efficiently approximate the sparse code $\mathbf{a}$ of input signal $\mathbf{x}$, which is obtained by solving (\ref{e000}) in advance. The LISTA network learns the hyperparameters as a general regression model from training data to their pre-solved sparse codes using back-propagation. 

LISTA overlooks the useful geometric information among data points \cite{zheng2011graph}, and therefore could be viewed as a \textbf{special case} of TAGnet in Fig. \ref{figISTA} when $\alpha$ = 0 (i.e., removing the $\times \mathbf{L}$ branches). Moreover, LISTA aims to approximate the ``optimal'' sparse codes pre-obtained from (\ref{e000}), and therefore requires the estimation of $\mathbf{D}$ and the tedious pre-computation of $\mathbf{A}$. The authors did not exploit its potential in supervised and task-specific feature learning.



\section{A Deeper Look: Hierarchical Clustering by DTAGnet}

Deep networks are well known for their capabilities to learn semantically rich representations by hidden layers \cite{DeCaf}. In this section, we investigate how the intermediate features $\mathbf{Z}_k$ ($k = 1, 2$) in TAGnet (Fig. \ref{figISTA} (a)) can be interpreted, and further utilized to improve the model, for specific clustering tasks. Compared to related non-deep models \cite{IJCAI}, such a hierarchical clustering property is another unique advantage of being deep. 

\begin{figure}[htbp]
\centering
\includegraphics[resolution=320]{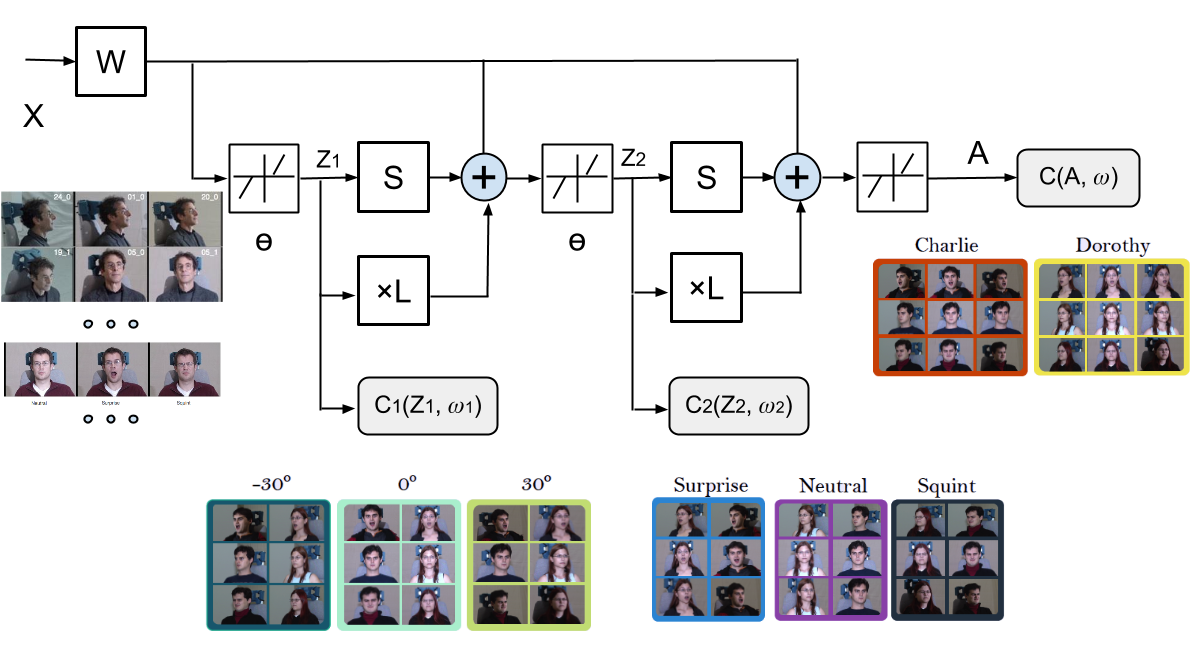}
\caption{The DTAGnet architecture, taking the CMU MultiPIE dataset as an example. The model is able to simultaneously learn features for pose clustering ($\mathbf{Z}_1$), for expression clustering ($\mathbf{Z}_2$), and for identity clustering ($\mathbf{A}$). The first two attributes are related to and helpful for the last (overall) task. Part of image sources are referred from \cite{MultiPIE} and \cite{JMLR}. }
\label{DSGLISTA}
\end{figure}

Our strategy is mainly inspired by the algorithmic framework of deeply supervised nets \cite{DSN}. As in Fig. \ref{DSGLISTA}, our proposed Deeply-Task-specific And Graph-regularized Network (\textbf{DTAGnet}) brings in additional deep feedbacks, by associating a clustering-oriented local auxiliary loss $C_k(\mathbf{Z}_k, \bm{\omega}_k)$ ($k$ = 1, 2) with each stage. Such an auxiliary loss takes the same form as the overall $C(\mathbf{A}, \bm{\omega})$, except that the expected cluster number may be different, depending on the auxiliary clustering task to be performed. The DTAGnet backpropagates errors not only from the overall loss layer, but also simultaneously from the auxiliary losses. 

While seeking the optimal performance of the target clustering, DTAGnet is also driven by two auxiliary tasks that are explicitly targeted at clustering specific attributes. It enforcrs constraint at each hidden representation for directly making a good cluster prediction. In addition to the overall loss, the introduction of auxiliary losses gives another strong push to obtain discriminative and sensible features at each individual stage. As discovered in the classification experiments in \cite{DSN}, the auxiliary loss both acts as feature regularization to reduce generalization errors and results in faster convergence.  We also find in Section V that every $\mathbf{Z}_k$ ($k = 1, 2$) is indeed most suited for its targeted task.

In \cite{JMLR}, a Deep Semi-NMF model was proposed to learn hidden representations, that grant themselves an interpretation of clustering according to different attributes. The authors considered the problem of mapping facial images to their identities. A face image also contains attributes like pose and expression that help identify the person depicted. In their experiments, the authors found that by further factorizing this mapping in a way that each factor adds an extra layer of abstraction, the deep model could automatically learn latent intermediate representations that are implied for clustering identity-related attributes. Although there is a clustering interpretation, those hidden representations are not specifically optimized in clustering sense. Instead, the entire model is trained with only the overall reconstruction loss, after which clustering is performed using K-means on learnt features. Consequently, their clustering performance is not satisfactory. Our study shares the similar observation and motivation with \cite{JMLR}, but in a more task-specific manner by performing the optimizations of auxiliary clustering tasks jointly with the overall task.

\section{Experiment Results}

\subsection{Datasets and measurements}
We evaluate the proposed model on three publicly available datasets:
\begin{itemize}
\item \textbf{MNIST} \cite{zheng2011graph} consists of a total number of 70, 000 quasi-binary, handwritten digit images, with digits 0 to 9. The digits are normalized and centered in fixed-size images of 28 $\times$ 28. 
\item \textbf{CMU MultiPIE} \cite{MultiPIE} contains around 750, 000 images of 337 subjects, that are captured under varied laboratory conditions. A unique property of CMU MultiPIE lies in that each image comes with labels for  the identity, illumination, pose and expression attributes. That is why CMU MultiPIE is chosen in \cite{JMLR} to learn multi-attribute features (Fig. \ref{DSGLISTA}) for hierarchical clustering. In our experiments, we follow \cite{JMLR} and adopt a subset of 13, 230 images of 147 subjects in 5 different poses and 6 different emotions. Notably, we do not pre-process the images by using piece-wise affine warping as utilized by \cite{JMLR} to align these images.
\item \textbf{COIL20} \cite{nene1996columbia} contains 1, 440 32 $\times$ 32 gray scale images of 20 objects (72 images per object). The images of each object were taken 5 degree apart. 
\end{itemize}
\begin{figure}[tbp]
\centering
\begin{minipage}{0.40\textwidth}
\centering \subfigure[] {
\includegraphics[width=\textwidth]{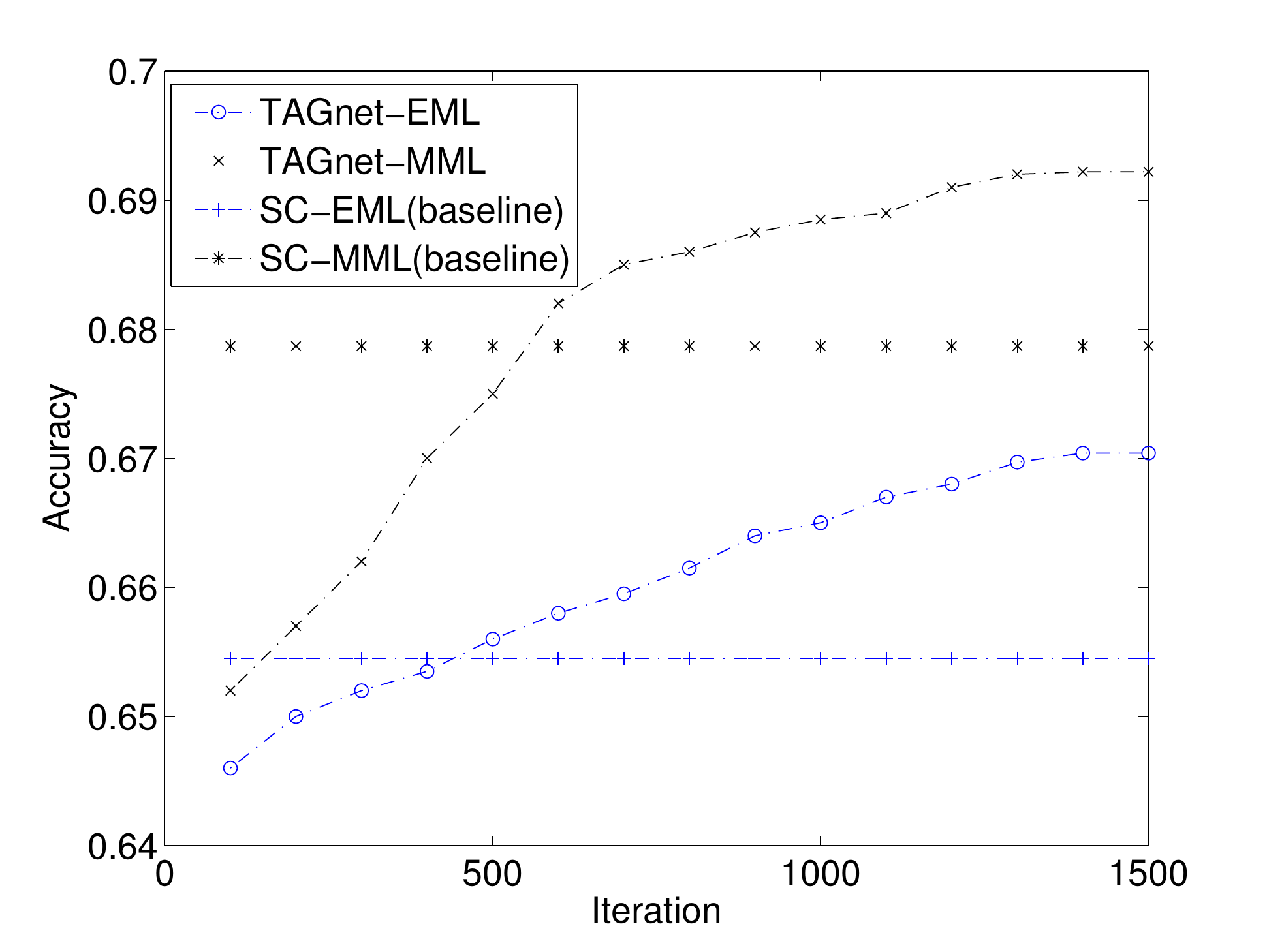}
}\\
\centering \subfigure[] {
\includegraphics[width=\textwidth]{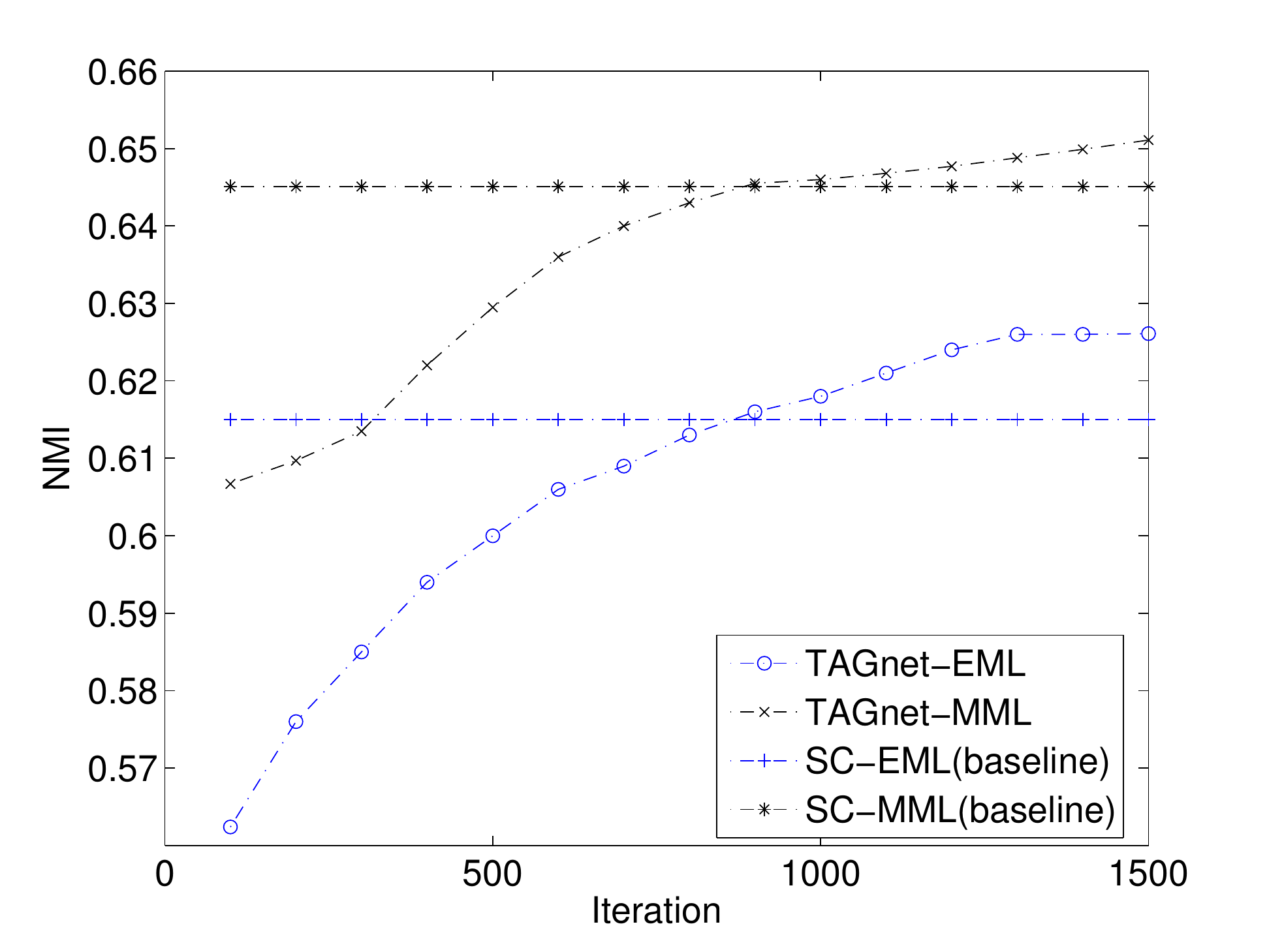}
}\end{minipage}
\caption{The accuracy and NMI plots of TAGnet-EML/TAGnet-MML on MNIST, starting from the initialization, and tested every 100 iterations. The accuracy and NMI of SC-EML/SC-MML are also plotted as baselines.}
\label{sccompare}
\end{figure}
Although the paper only evaluates the proposed method using image datasets, the methodology itself is not limited to only image subjects. We apply two widely-used measures to evaluate the clustering performances: the accuracy and the Normalized Mutual Information(NMI) \cite{zheng2011graph}, \cite{cheng2010learning}. We follow the convention of many clustering work \cite{zheng2011graph, yang2014data, IJCAI}, and do not distinguish training from testing. We train our models on all available samples of each dataset, reporting the clustering performances as our testing results. Results are averaged from 5 independent runs.


\subsection{Experiment settings}

The proposed networks are implemented using the cuda-convnet package \cite{ImageNet}. The network takes $K$ = 2 stages by default. We apply a constant learning rate of 0.01 with no momentum to all trainable layers. The batch size of 128. In particular, to encode graph regularization as a prior, we fix $\mathbf{L}$ during model training by setting its learning rate to be 0. Experiments run on a workstation with 12 Intel Xeon 2.67GHz CPUs and 1 GTX680 GPU. The training takes approximately 1 hour on the MNIST dataset. It is also observed that the training efficiency of our model scales approximately linearly with data. 


In our experiments, we set the default value of $\alpha$ to be 5, $p$ to be 128, and $\lambda$ to be chosen from [0.1, 1] by cross-validation\footnote{The default values of $\alpha$ and $p$ are inferred from the related sparse coding literature\cite{zheng2011graph}, and validated in experiments.}. A dictionary $\mathbf{D}$ is first learned from $\mathbf{X}$ by K-SVD \cite{elad2006image}. $\mathbf{W}$,  $\mathbf{S}$ and $\bm{\theta}$ are then initialized based on (\ref{scpara}). $\mathbf{L}$ is also pre-calculated from $\mathbf{P}$, which is formulated by the Gaussian Kernel: $\mathbf{P}_{ij} = \exp(-\frac{||\mathbf{x}_i - \mathbf{x}_j||^2_2}{\delta^2})$ ($\delta$ is also selected by cross-validation). After obtaining the output $\mathbf{A}$ from the initial (D)TAGnet models, $\bm{\omega}$ (or  $\bm{\omega}_k$) could be initialized based on minimizing (\ref{logistic}) or (\ref{hingeall}) over $\mathbf{A}$ (or $\mathbf{Z}_k$).

\subsection{Comparison experiments and analysis}


\begin{figure*}[htbp]
\centering
\begin{minipage}{0.33\textwidth}
\centering \subfigure[] {
\includegraphics[width=\textwidth]{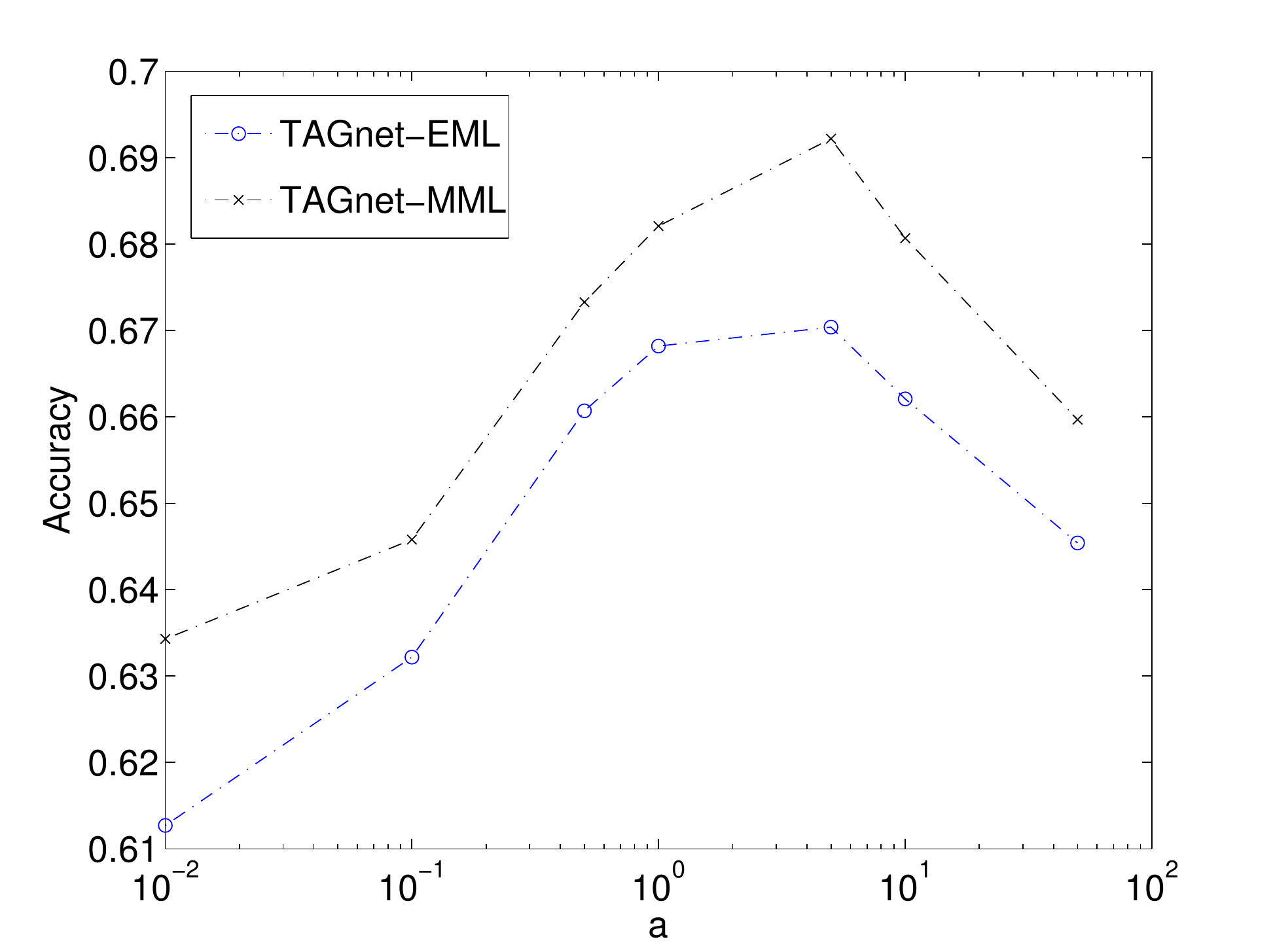}
}\\
\centering \subfigure[] {
\includegraphics[width=\textwidth]{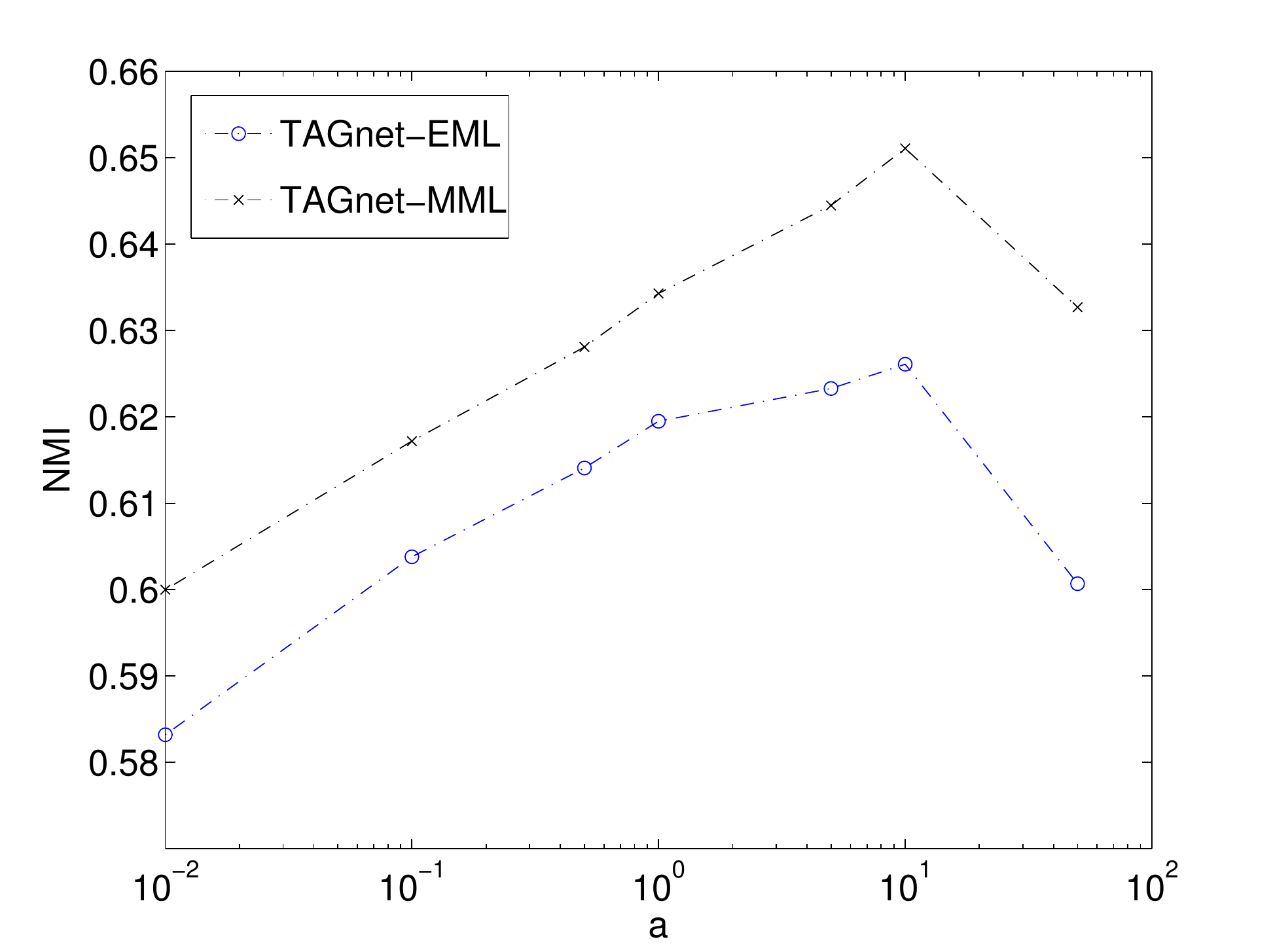}
}\end{minipage}
\begin{minipage}{0.33\textwidth}
\centering \subfigure[] {
\includegraphics[width=\textwidth]{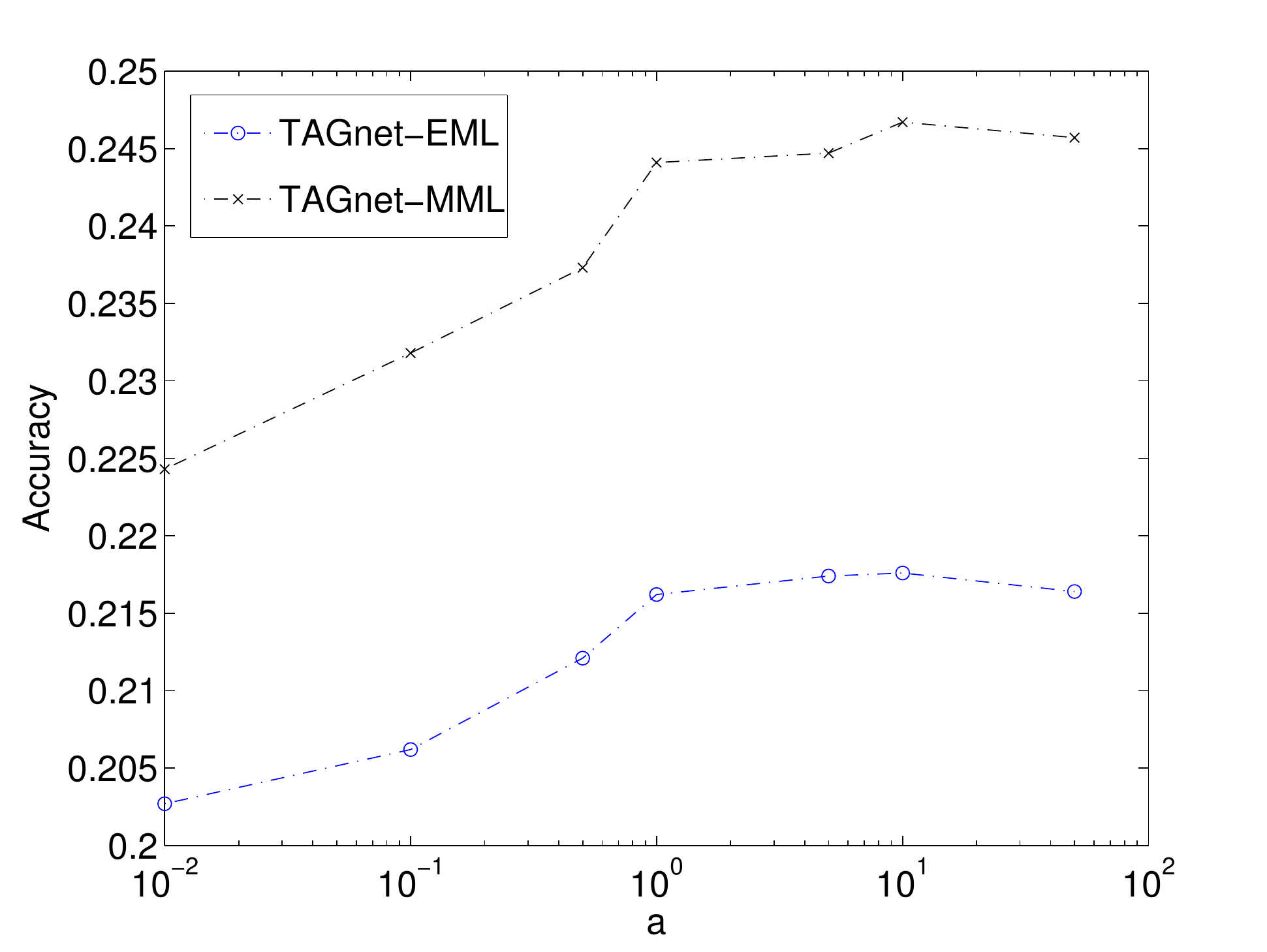}
}\\
\centering \subfigure[] {
\includegraphics[width=\textwidth]{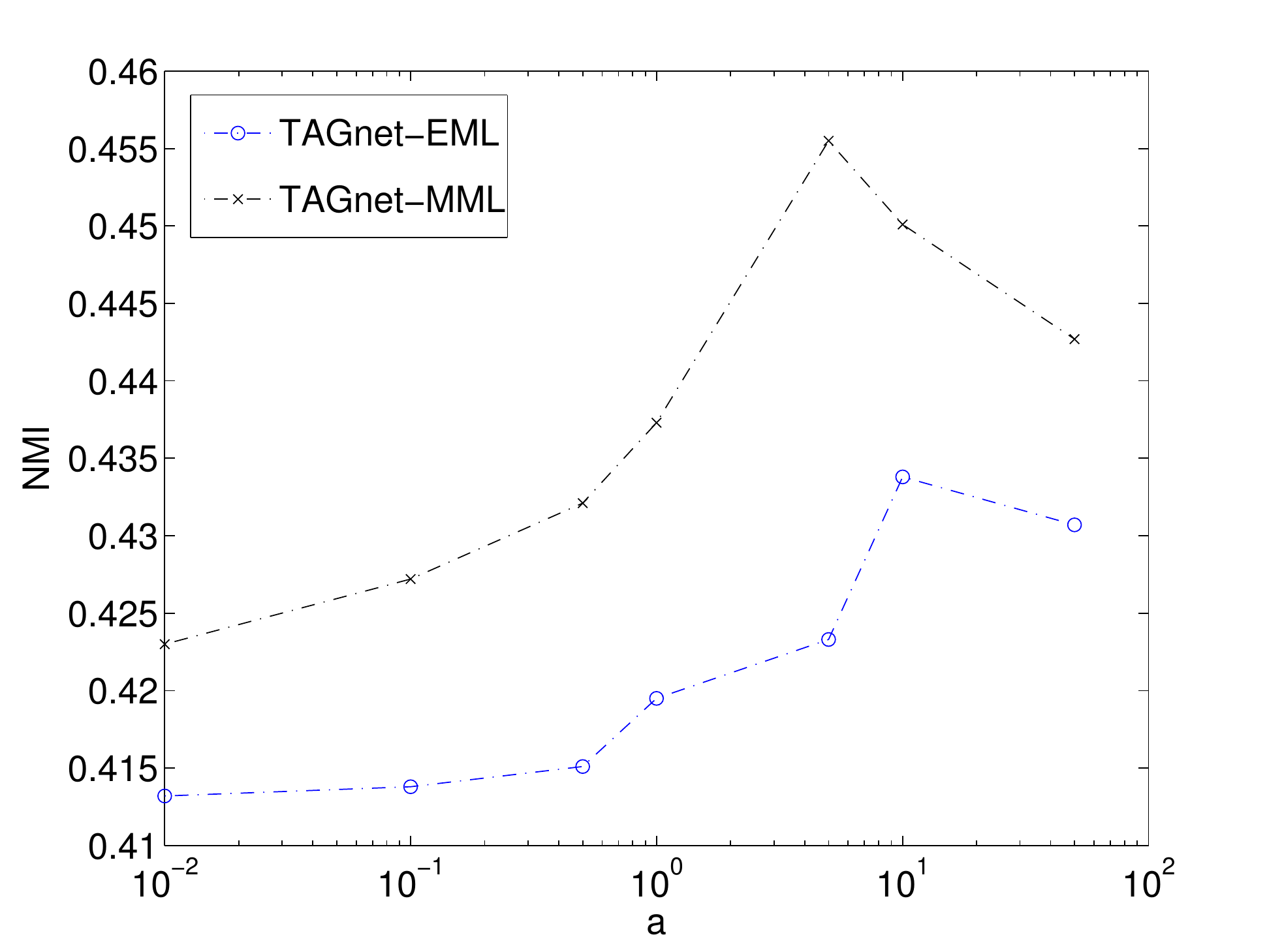}
}\end{minipage}
\begin{minipage}{0.33\textwidth}
\centering \subfigure[] {
\includegraphics[width=\textwidth]{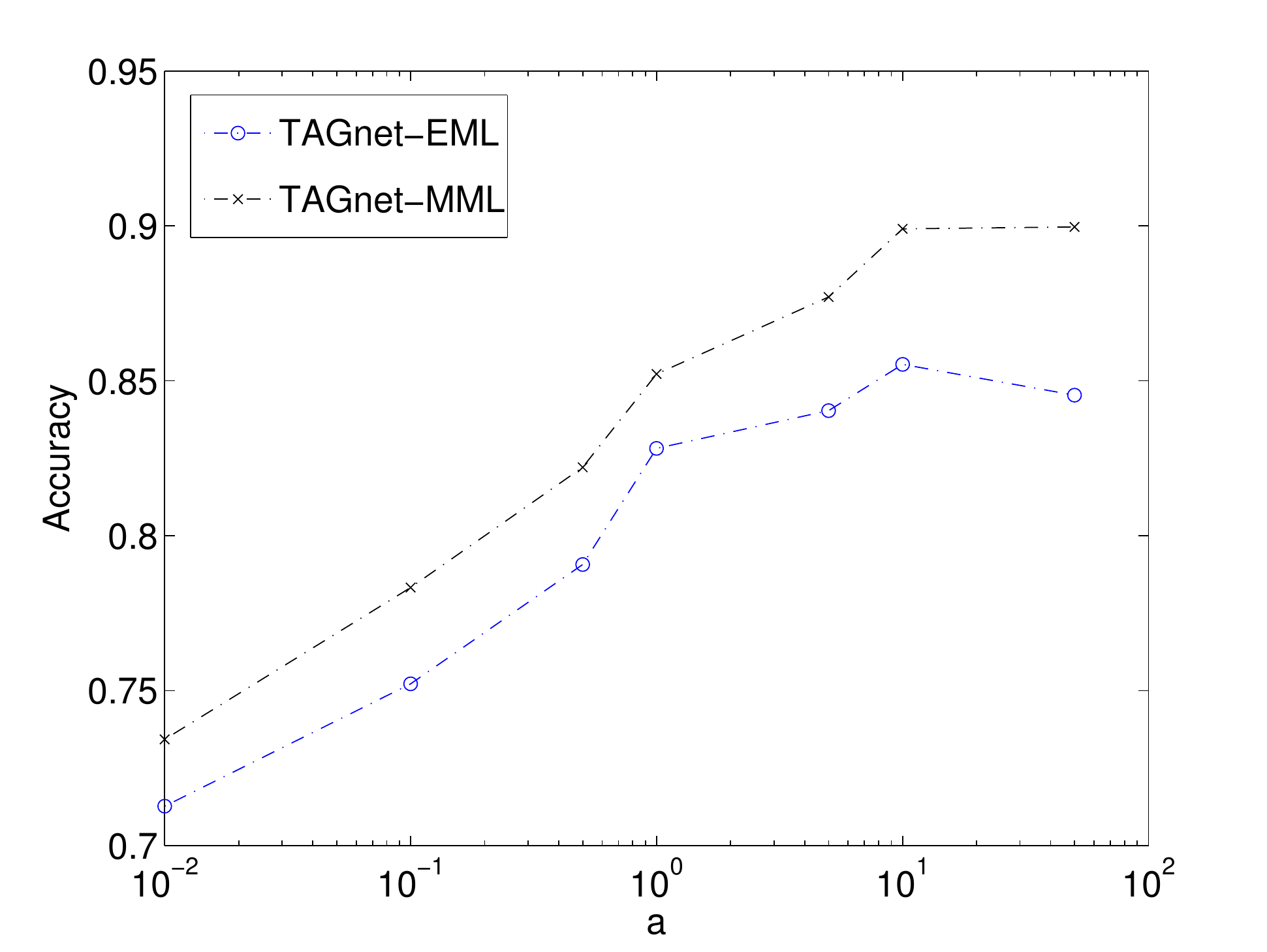}
}\\
\centering \subfigure[] {
\includegraphics[width=\textwidth]{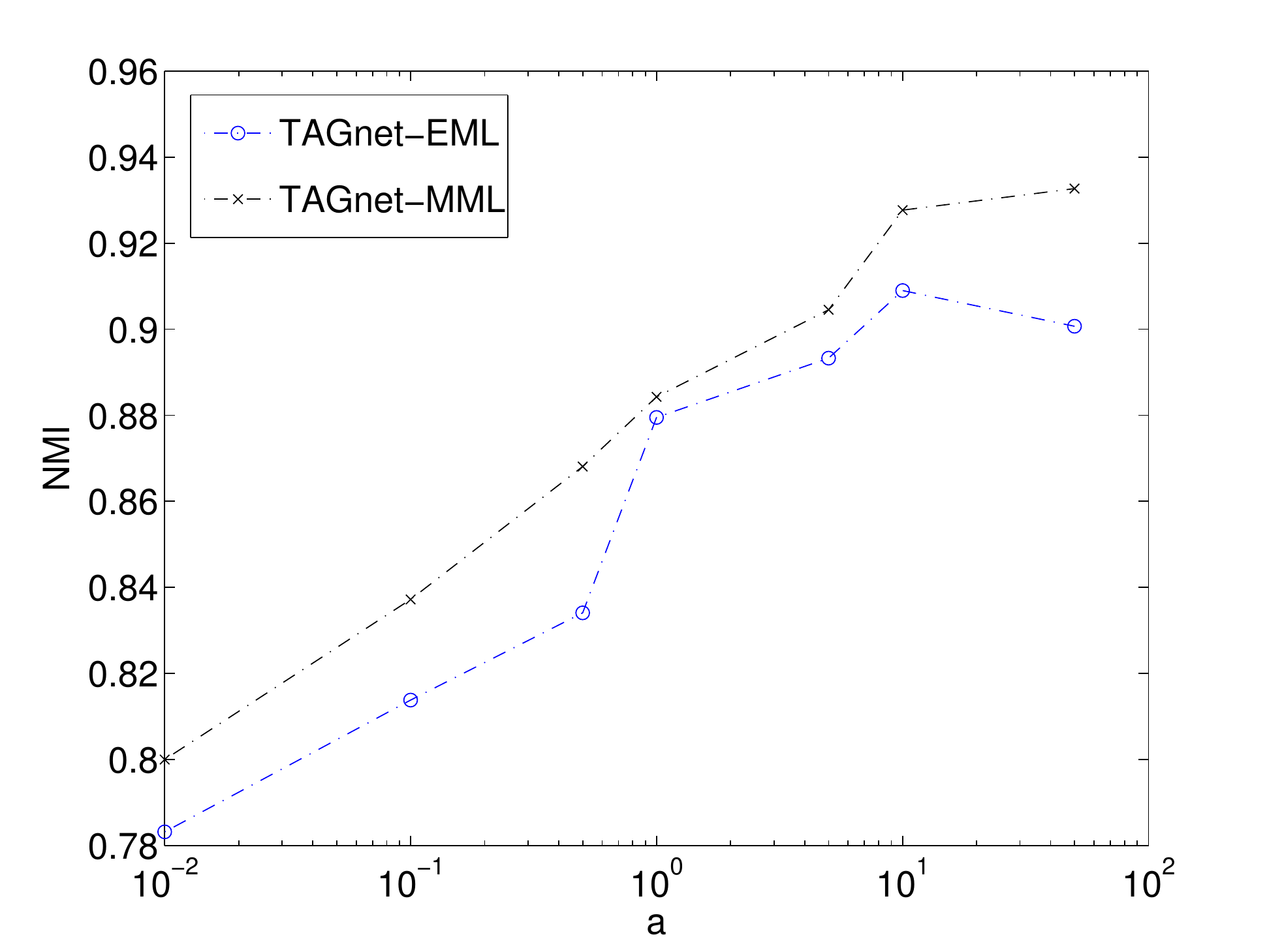}
}\end{minipage}
\caption{The clustering accuracy and NMI plots (x-axis logarithm scale) of TAGnet-EML/TAGnet-MML versus the parameter choices of $\alpha$, on: (a) (b) MNIST; (c) (d) CMU MultiPIE; (e) (f) COIL20.}
\label{alpha}
\end{figure*}

\subsubsection{Benefits of the task-specific deep architecture}
We denote the proposed model of TAGnet plus entropy-minimization loss (EML) (\ref{logistic}) as TAGnet-EML, and the one plus maximum-margin loss (MML) (\ref{hingeall}) as TAGnet-MML, respectively. We include the following comparison methods:
\begin{itemize}
\item We refer to the initializations of the proposed joint models as their \textbf{``Non-Joint'' }counterparts, denoted as NJ-TAGnet-EML and NJ-TAGnet-MML (NJ short for non-joint), respectively. 
\item We design a \textbf{Baseline Encoder} (BE), which is a fully-connected feedforward network, consisting of three hidden layers of dimension $p$ with ReLU neuron. It is obvious that the BE has the \textit{same parameter complexity} as TAGnet\footnote{except for the ``$\theta$" layers, each of which contains only $p$ free parameters and thus ignored}. The BEs are also tuned by EML or MML in the same way, denoted as BE-EML or BE-MML, respectively. We intend to verify our important argument, that \textit{the proposed model benefits from the task-specific TAGnet architecture, rather than just the large learning capacity of generic deep models}.
\item We compare the proposed models with their closest \textbf{``shallow''} competitors, i.e., the joint optimization methods of graph-regularized sparse coding and discriminative clustering in \cite{IJCAI}. We re-implement their work using both (\ref{logistic}) or (\ref{hingeall}) losses, denoted as SC-EML and SC-MML (SC short for sparse coding). Since in \cite{IJCAI} the authors already revealed SC-MML outperforms the classical methods such as MMC and $\ell_1$ graph methods, we do not compare with them again. 
\item We also include \textbf{Deep Semi-NMF} \cite{JMLR} as a state-of-the-art deep learning-based clustering work. We mainly compare our results with their reported performances on CMU MultiPIE. \footnote{With various component numbers tested in \cite{JMLR}, we choose their best cases (60 components).}.
\end{itemize}




As revealed by the full comparison results in Table \ref{results}, the proposed task-specific deep architectures outperform other with a noticeable margin. The underlying domain expertise guides the data-driven training in a more principled way. In contrast, the ``general-architecture'' baseline encoders (BE-EML and BE-MML) appear to produce much worse (even worst) results. Furthermore, it is evident that the proposed end-to-end optimized models outperform their ``non-joint'' counterparts. For example, on the MNIST dataset,TAGnet-MML surpasses NJ-TAGnet-MML by around 4\% in accuracy and 5\% in NMI. 

By comparing the TAGnet-EML/TAGnet-MML with SC-EML/SC-MML, we draw a promising conclusion:  adopting a more parameterized deep architecture allows a larger feature learning capacity compared to conventional sparse coding. Although similar points are well made in many other fields \cite{ImageNet}, we are interested in a closer look between the two. Fig. \ref{sccompare} plots the clustering accuracy and NMI curves of TAGnet-EML/TAGnet-MML on the MNIST dataset, along with iteration numbers. Each model is well initialized at the very beginning, and the clustering accuracy and NMI are computed every 100 iterations. At first, the clustering performances of deep models are even slightly worse than sparse-coding methods, mainly since the initialization of TAGnet hinges on a truncated approximated of graph-regularized sparse coding. After a small number of iterations, the performance of the deep models surpass sparse coding ones, and continue rising monotonically until reaching a higher plateau.

 \begin{table*}[t]
 \scriptsize
\begin{center}
\caption{Accuracy and NMI performance comparisons on all three datasets}
\label{results}
\begin{tabular}{|c|c|c|c|c|c|c|c|c|c|c|}
\hline
 & & TAGnet & TAGnet  & NJ-TAGnet & NJ-TAGnet & BE  & BE & SC  & SC & Deep \\
 & & -EML & -MML  & -EML & -MML & -EML  & -MML & -EML  & -MML & Semi-NMF\\
 \hline
$\multirow{2}{*}{\textit{MNIST}}$ &Acc & 0.6704 & 0.6922 & 0.6472 & 0.5052 & 0.5401 & 0.6521 & 0.6550 & 0.6784  & $\slash$ \\
\cline{2-11}
$ $ &NMI  & 0.6261 & 0.6511 &0.5624 & 0.6067  & 0.5002 & 0.5011 & 0.6150 & 0.6451 & $\slash$\\
\hline
\textit{CMU} &Acc & 0.2176 & 0.2347 & 0.1727 & 0.1861 & 0.1204 & 0.1451 & 0.2002  & 0.2090  & 0.17 \\
\cline{2-11}
\textit{MultiPIE}  &NMI  & 0.4338 & 0.4555 & 0.3167 & 0.3284 & 0.2672 & 0.2821 & 0.3337  & 0.3521 & 0.36 \\
\hline
$\multirow{2}{*}{\textit{COIL20}}$ &Acc & 0.8553 & 0.8991 & 0.7432 & 0.7882  & 0.7441 & 0.7645 & 0.8225 & 0.8658 & $\slash$ \\
\cline{2-11}
$$ &NMI  & 0.9090 & 0.9277 & 0.8707  & 0.8814  & 0.8028 & 0.8321 & 0.8850 & 0.9127  & $\slash$\\
\hline
\end{tabular}
\end{center}
\end{table*}

\subsubsection{Effects of graph regularization}

In (\ref{e00}), the graph regularization term imposes stronger smoothness constraints on the sparse codes with a larger $\alpha$. It also happens to the TAGnet. We investigate how the clustering performances of TAGnet-EML/TAGnet-MML are influenced by various $\alpha$ values. From Fig. \ref{alpha}, we observe the identical general tendency on all three datasets. While $\alpha$ increases, the accuracy/NMI result will first rise then decrease, with the peak appearing between $\alpha \in$ [5, 10]. As an interpretation, the local manifold information is not sufficiently encoded when $\alpha$ is too small ($\alpha$ = 0 will completely disable the $\times \mathbf{L}$ branch of TAGnet, and reduces its to the LISTA network \cite{LISTA} fine-tuned by the losses). On the other hand, when $\alpha$ is large, the sparse codes are ``over-smoothened'' with a reduced discriminative ability. Note that similar phenomenons are also reported in other relevant literature, e. g. , \cite{zheng2011graph, IJCAI}.

%
\begin{figure}[htbp]
\centering
\begin{minipage}{0.40\textwidth}
\centering \subfigure[] {
\includegraphics[width=\textwidth]{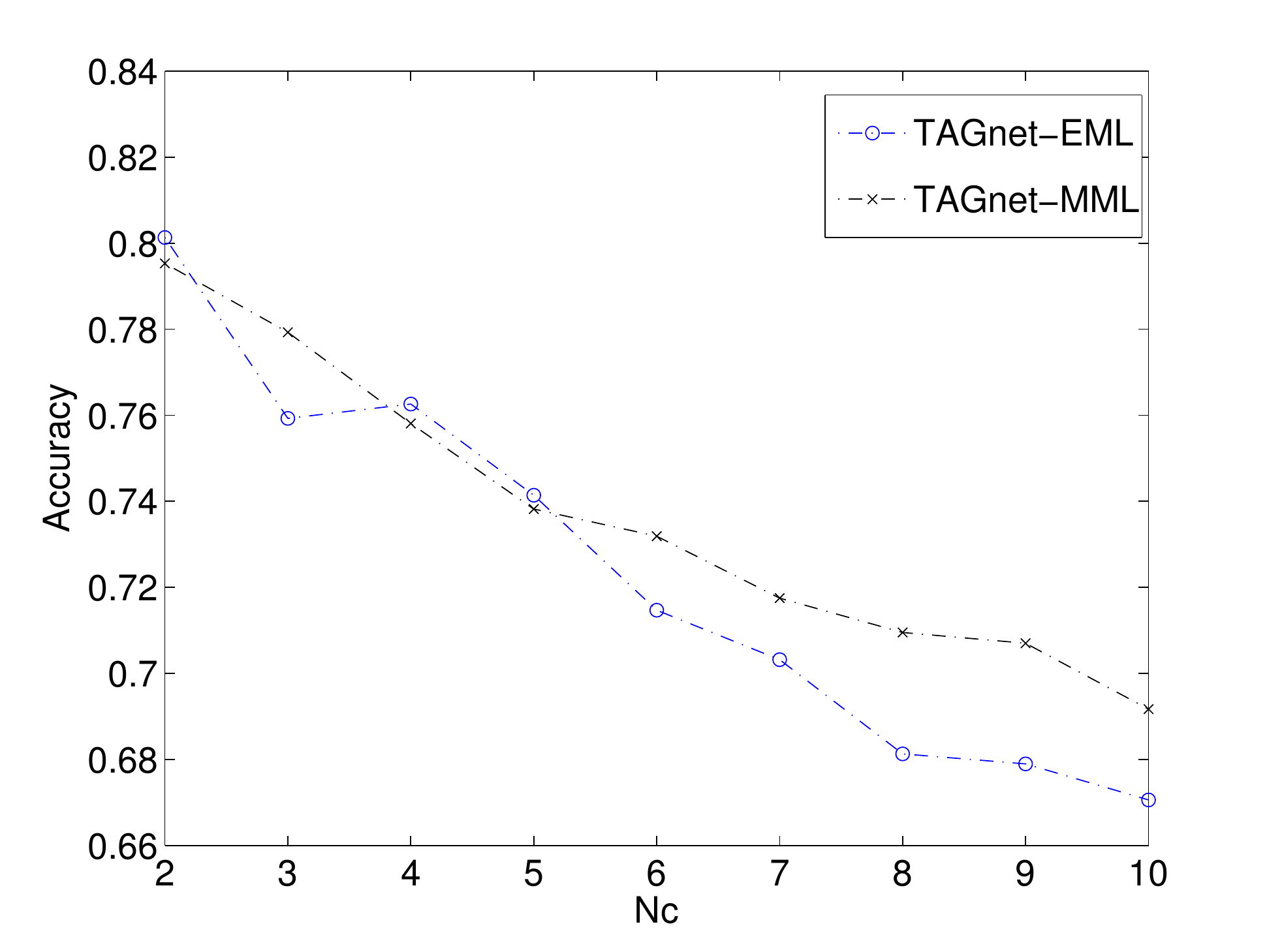}
}\\
\centering \subfigure[] {
\includegraphics[width=\textwidth]{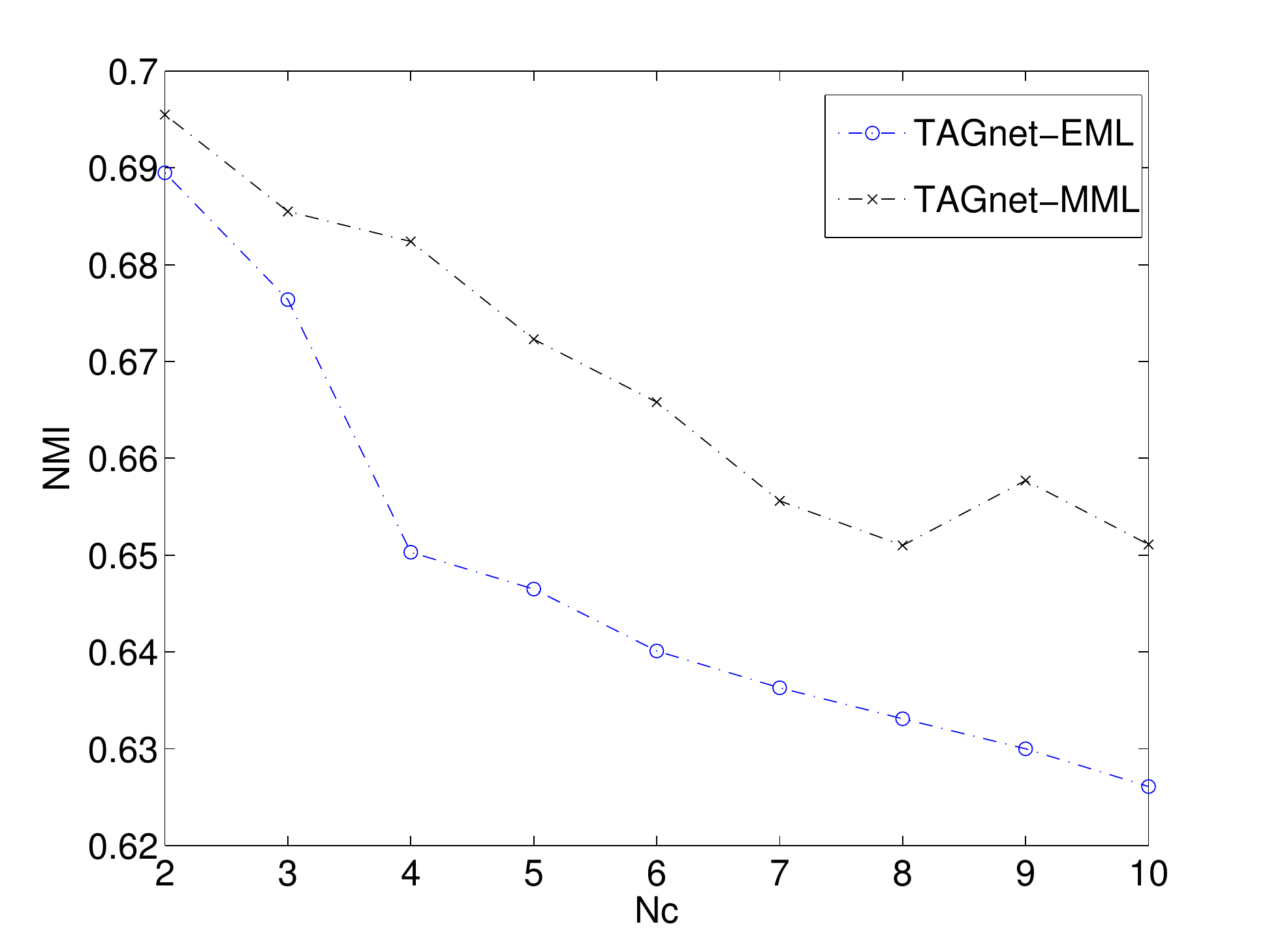}
}\end{minipage}
\caption{The clustering accuracy and NMI plots of TAGnet-EML/TAGnet-EML versus the cluster number $N_c$ ranging from 2 to 10, on MNIST.}
\label{time}
\end{figure}

Furthermore, comparing among Fig. \ref{alpha} (a) - (f), it is noteworthy to observe how graph regularization behaves differently on three of them. We notice that the COIL20 dataset is the one that is the most sensitive to the choice of $\alpha$. Increasing $\alpha$ from 0.01 to 50 leads to a improvement of more than 10\%, in terms of both accuracy and NMI. It verifies the significance of graph regularization when trying samples are limited \cite{yang2014data}. On the MNIST dataset, both models obtain a gain of up to 6\% in accuracy and 5\% in NMI, by tuning $\alpha$ from 0.01 to 10. However, unlike COIL20 that almost always favors larger $\alpha$, the model performance on the MNIST dataset tends to be not only saturated, but even significantly hampered when $\alpha$ continues rising to 50. The CMU MultiPIE dataset witnesses moderate improvements of around 2\% in both measurements. It is not as sensitive to $\alpha$ as the other two. Potentially, it might be due to the complex variability in original images that makes the graph $\mathbf{W}$ unreliable for estimating the underlying manifold geometry. We suspect that more sophisticated graphs may help alleviate the problem, and will explore it in future. 


\subsubsection{Scalability and robustness} 

On the MNIST dataset, We re-conduct the clustering experiments with the cluster number $N_c$ ranging from 2 to 10, using TAGnet-EML/TAGnet-MML.  Fig. \ref{time} shows that the clustering accuracy and NMI change by varying the number of clusters. The clustering performance transits smoothly and robustly when the task scale changes.

\begin{figure}[htbp]
\begin{minipage}{0.40\textwidth}
\centering \subfigure[] {
\includegraphics[width=\textwidth]{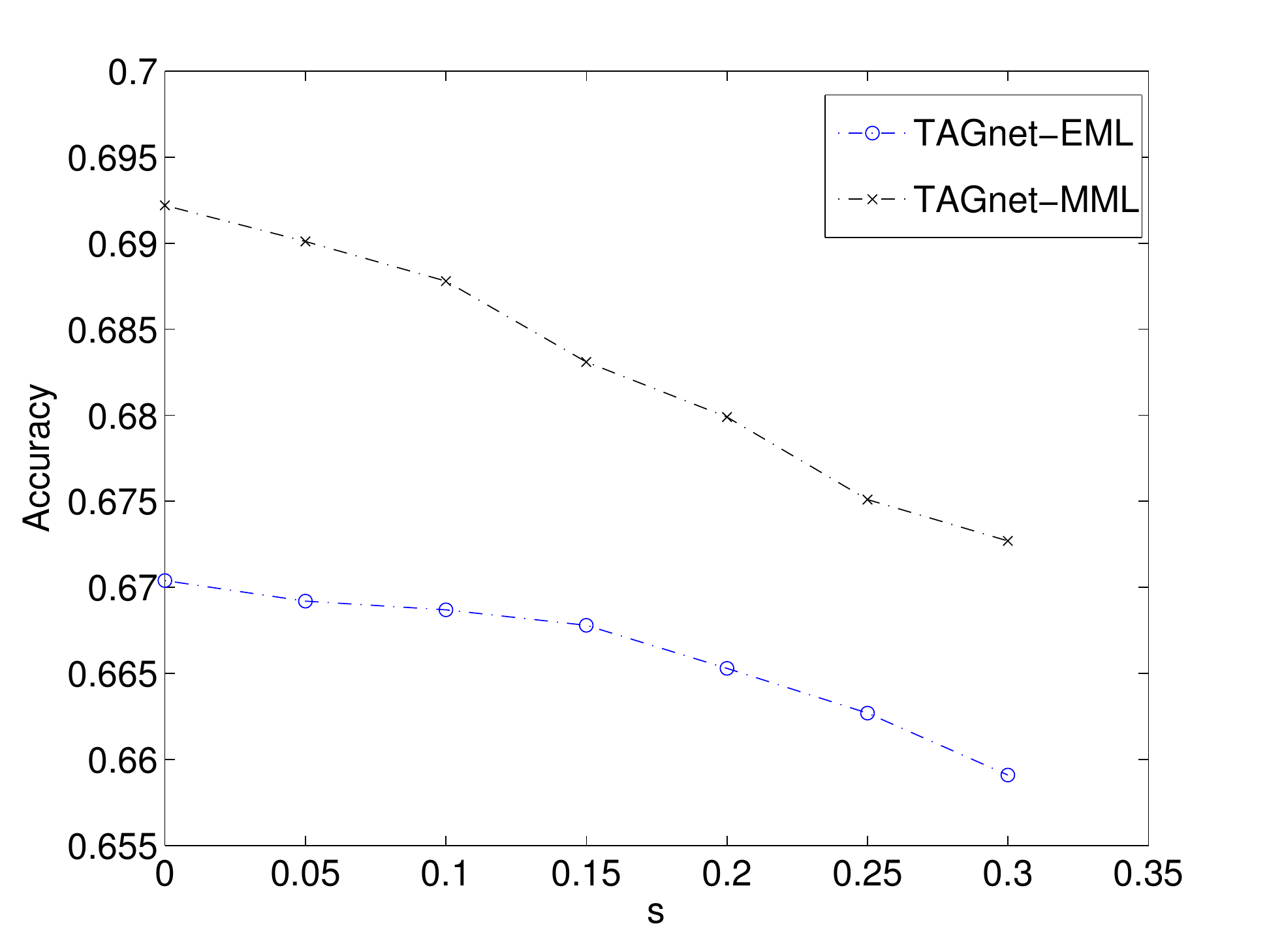}
}\\
\centering \subfigure[] {
\includegraphics[width=\textwidth]{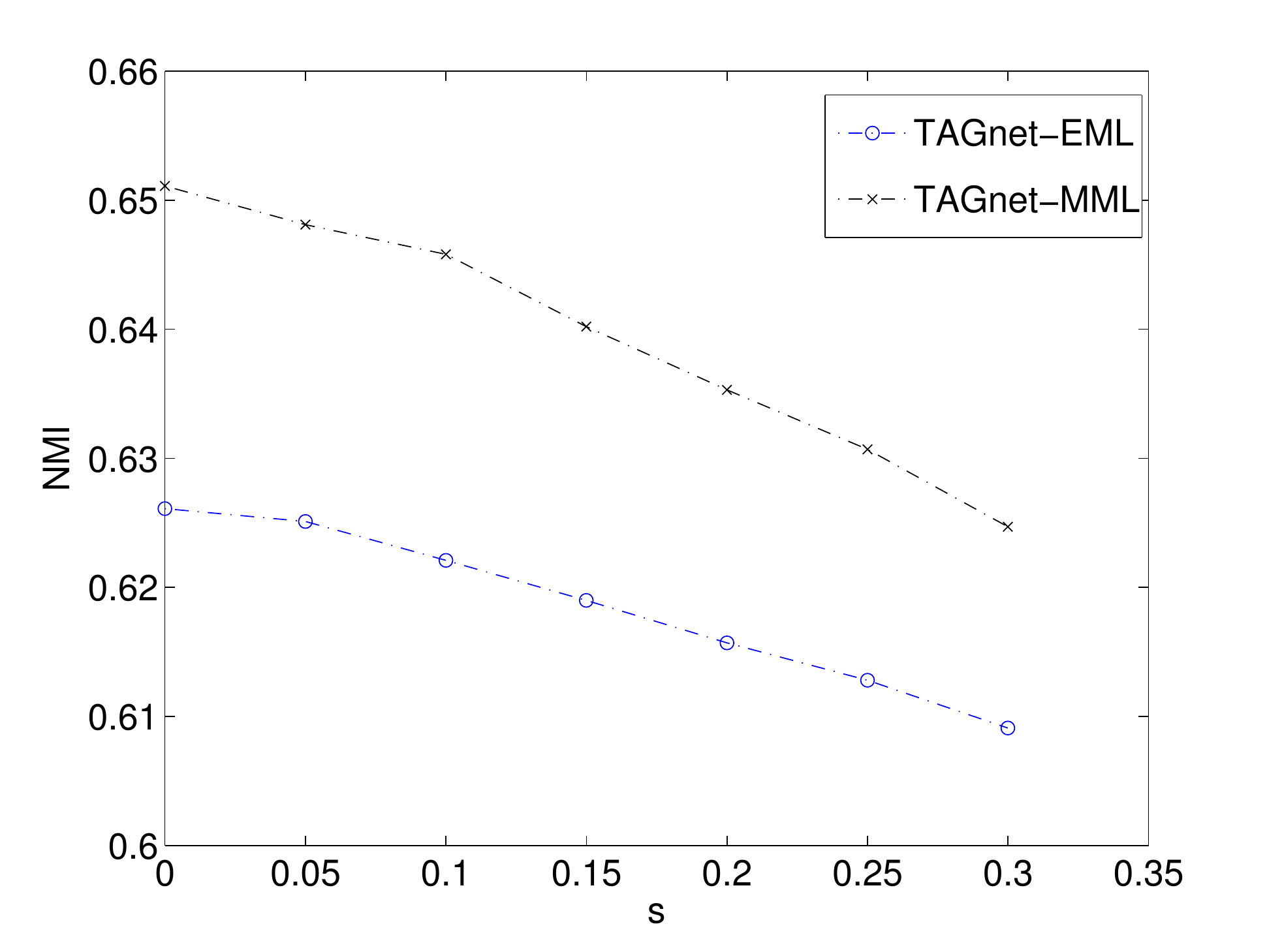}
}\end{minipage}
\caption{The clustering accuracy and NMI plots of TAGnet-EML/TAGnet-MML versus the noise level $s$, on MNIST.}
\label{noise}
\end{figure}

To examine the proposed models' robustness to noise, we add various Gaussian noise, whose standard deviation $s$ ranges from 0 (noiseless) to 0.3, to re-train our MNIST model. Fig. \ref{noise} indicates that both TAGnet-EML and TAGnet-MML own certain robustness to noise. When $s$ is less than 0.1, there is even little visible performance degradation. While TAGnet-MML constantly outperforms TAGnet-EML in all experiments (as MMC is well-known to be highly discriminative \cite{xu2004maximum} ), it is interesting to observe in Fig. \ref{noise} that the latter one is slightly more robust to noise than the former. It is perhaps owing to the probability-driven loss form (\ref{logistic}) of EML that allows for more flexibility.

\subsection{Hierarchical clustering on CMU MultiPIE}

As observed, CMU MultiPIE is very challenging for the basic identity clustering task. However, it comes with several other attributes: pose, expression, and illumination, which could be of assistance in our proposed DTAGnet framework. In this section, we apply the similar setting of \cite{JMLR} on the same CMU MultiPIE subset, by setting pose clustering as the Stage I auxiliary task, and expression clustering as the Stage II auxiliary task\footnote{In fact, although claimed to be applicable to multiple attributes, \cite{JMLR} only examined the first level features for pose clustering without considering expressions, since it relied on a warping technique to pre-process images, that gets rid of most expression variability. }. In that way, we target $C_1(\mathbf{Z}_1, \bm{\omega}_1)$ at 5 clusters,  $C_2(\mathbf{Z}_2, \bm{\omega}_2)$ at 6 clusters, and finally $C(\mathbf{A}, \bm{\omega})$ as 147 clusters. 

The training of DTAGnet-EML/DTAGnet-MML follows the same aforementioned process except for considering extra back-propagated gradients from task $C_k(\mathbf{Z}_k, \bm{\omega}_k)$ in Stage $k$ ($k$ = 1, 2). After then, we test each $C_k(\mathbf{Z}_k, \bm{\omega}_k)$ separately on their targeted task. In DTAGnet, each auxiliary task is also jointly optimized with its intermediate feature $\mathbf{Z}_k$, which differentiate our methodology substantially from \cite{JMLR}. It is thus no surprise to see in Table \ref{confusion} that each auxiliary task obtains much improved performances than \cite{JMLR}\footnote{In \cite{JMLR} Table. 2, it reports that the best accuracy of pose clustering task falls around 28\%, using the most suited layer features. } Most notably, the performances of the overall identity clustering task witness \textbf{a very impressive boost of around 7\% in accuracy}. We also test DTAGnet-EML/DTAGnet-MML with only $C_1(\mathbf{Z}_1, \bm{\omega}_1)$ or $C_2(\mathbf{Z}_2, \bm{\omega}_2)$ kept. Experiments verify that by adding auxiliary tasks gradually, the overall task keeps being benefited. Those auxiliary tasks, when enforced together, can also reinforce each other mutually. 

 \begin{table}[htbp]
 \begin{center}
 \caption{Effects of incorporating auxiliary clustering tasks in DTAGnet-EML/DTAGnet-MML (P: Pose; E: Expression; I: Identity)}
 \label{confusion}
 \begin{tabular}{|c|c|c|c|c|c|c|}
 \hline
$\multirow{2}{*}{Method}$  & \multicolumn{2}{c|}{Stage I} & \multicolumn{2}{c|}{Stage II } & \multicolumn{2}{c|}{Overall} \\
\cline{2-7}
$$ & Task & Acc & Task & Acc & Task & Acc \\
\hline
$\multirow{4}{*}{DTAGnet}$ &  $\slash$ & $\slash$ & $\slash$ & $\slash$ &  I &  0.2176 \\
\cline{2-7}
$$ & P &  0.5067 & $\slash$ & $\slash$&  I & 0.2303 \\
\cline{2-7}
$\multirow{2}{*}{-EML}$ &  $\slash$ & $\slash$ & E & 0.3676 &  I &  0.2507 \\
\cline{2-7}
$$ & P &  0.5407 & E & 0.4027 &  I & 0.2833 \\
 \hline
 \hline
 $\multirow{4}{*}{DTAGnet}$ &  $\slash$ & $\slash$ & $\slash$ & $\slash$ &  I &  0.2347 \\
\cline{2-7}
$$ & P &  0.5251 & $\slash$ & $\slash$&  I & 0.2635 \\
\cline{2-7}
$\multirow{2}{*}{-MML}$  &  $\slash$ & $\slash$ & E & 0.3988 &  I &  0.2858 \\
\cline{2-7}
$$ & P &  0.5538 & E & 0.4231 &  I & 0.3021 \\
 \hline
 \end{tabular}
 \end{center}
 \end{table}

One might be curious that, \textbf{which one matters more in the performance boost}: the deeply task-specific architecture that brings extra discriminative feature learning, or the proper design of auxiliary tasks that capture the intrinsic data structure characterized by attributes? 

 \begin{table}[htbp]
 \begin{center}
 \caption{Effects of varying target cluster numbers of auxiliary tasks in DTAGnet-EML/DTAGnet-MML}
 \label{confusion2}
 \begin{tabular}{|c|c|c|c|}
 \hline
$\multirow{2}{*}{Method}$   & \#clusters& \#clusters & Overall\\
$$ &  in Stage I  &  in Stage II & Accuracy \\ 
\hline
$\multirow{4}{*}{DTAGnet}$ &  4  & 4 & 0.2827 \\
\cline{2-4}
$$ & 8  & 8 &  0.2813 \\
\cline{2-4}
$\multirow{2}{*}{-EML}$ &  12 & 12 & 0.2802 \\
\cline{2-4}
$$ & 20 &  20 & 0.2757 \\
 \hline
 \hline
$\multirow{4}{*}{DTAGnet}$ &  4  & 4 & 0.3030 \\
\cline{2-4}
$ $ & 8  & 8 &  0.3006 \\
\cline{2-4}
$\multirow{2}{*}{-MML}$ &  12 & 12 & 0.2927 \\
\cline{2-4}
$ $ & 20 &  20 & 0.2805 \\
 \hline
 \end{tabular}
 \end{center}
 \end{table}

To answer this important question, we vary the target cluster number in either $C_1(\mathbf{Z}_1, \bm{\omega}_1)$ or $C_2(\mathbf{Z}_2, \bm{\omega}_2)$, and re-conduct the experiments. Table \ref{confusion2} reveals that more auxiliary tasks, even those without any striaghtforward task-specific interpretation (e.g., partitioning the Multi-PIE subset into 4, 8, 12 or 20 clusters hardly makes semantic sense), may still help gain better performances. It is comprehensible that they simply promote more discriminative feature learning in a low-to-high, coarse-to-fine scheme. In fact, it is a complementary observation to the conclusion found in classification \cite{DSN}. On the other hand, at least in this specific case, while the target cluster numbers of auxiliary tasks get closer to the ground-truth (5 and 6 here), the models seem to achieve the best performances. We conjecture that when properly ``matched'' , every hidden representation in each layer is in fact most suited for clustering the attributes corresponding to the layer of interest. The whole model can be resembled to the problem of sharing low-level feature filters among several relevant high-level tasks in convolutional networks \cite{glorot2011domain}, but in a distinct context. 

We hence conclude that, the deeply-supervised fashion shows to be helpful for the deep clustering models, even when there are no explicit attributes for constructing a practically meaningful hierarchical clustering problem. However, it is preferable to exploit those attributes when available, as they lead to not only superior performances but more clearly interpretable models. The learned intermediate features can be potentially utilized for multi-task learning \cite{Qing}. 

\section{Conclusion}
In this paper, we present a deep learning-based clustering framework. Trained from end to end, it features a task-specific deep architecture inspired by the sparse coding domain expertise, which is then optimized under clustering-oriented losses. Such a well-designed architecture leads to more effective initialization and training, and significantly outperforms generic architectures of the same parameter complexity. The model could be further interpreted and enhanced, by introducing auxiliary clustering losses to the intermediate features. Extensive experiments verify the effectiveness and robustness of the proposed models.

\bibliographystyle{ieee}
\bibliography{icdm}

\end{document}